\documentclass[journal ]{new-aiaa}
\usepackage[utf8]{inputenc}
\usepackage{textcomp}

\usepackage{graphicx}
\usepackage{amsmath}
\usepackage[version=4]{mhchem}
\usepackage{siunitx}
\usepackage{longtable,tabularx}
\usepackage{algorithm} 
\usepackage[noend]{algpseudocode} 
\usepackage{lipsum}
\usepackage{subcaption}

\usepackage{booktabs}
\newcommand{\ra}[1]{\renewcommand{\arraystretch}{#1}}

\title{Satellite Chasers: Divergent Adversarial Reinforcement Learning to Engage Intelligent Adversaries on Orbit}

\author{Cameron Mehlman \footnote{PhD Candidate, Sibley School of Mechanical and Aerospace Engineering, 124 Hoy Rd., Ithaca NY 14850,  \\ \null\quad\quad (\textit{Corresponding Author : cpm222@cornell.edu}).} and Gregory Falco \footnote{Assistant Professor, Sibley School of Mechanical and Aerospace Engineering, 124 Hoy Rd., Ithaca NY 14850.}}
\affil{Cornell University, Ithaca, NY, 14850}

\begin{document}

\maketitle

\begin{abstract}
As space becomes increasingly crowded and contested, robust autonomous capabilities for multi-agent environments are gaining critical importance. Current autonomous systems in space primarily rely on optimization-based path planning or long-range orbital maneuvers, which have not yet proven effective in adversarial scenarios where one satellite is actively pursuing another. We introduce Divergent Adversarial Reinforcement Learning (DARL), a two-stage Multi-Agent Reinforcement Learning (MARL) approach designed to train autonomous evasion strategies for satellites engaged with multiple adversarial spacecraft. Our method enhances exploration during training by promoting diverse adversarial strategies, leading to more robust and adaptable evader models. We validate DARL through a cat-and-mouse satellite scenario, modeled as a partially observable multi-agent capture the flag game where two adversarial `cat' spacecraft pursue a single `mouse' evader.  DARL’s performance is compared against several benchmarks, including an optimization-based satellite path planner, demonstrating its ability to produce highly robust models for adversarial multi-agent space environments.
\end{abstract}

\section*{Nomenclature}

{\renewcommand\arraystretch{1.0}
\noindent\begin{longtable*}{@{}l @{\quad=\quad} l@{}}

$\boldsymbol{a}_e$ & action in Newtons \\

$c_{KL}$ & divergent loss constant \\

$D_{KL}$ & Kullback-Leibler divergence \\

$\boldsymbol{g}$ & goal position in meters \\

$\boldsymbol{H}_f$ & Histogram matrix \\

$n$ & orbit rate in $s^{-1}$ \\

$\boldsymbol{p}$ & position in cartesian coordinates in meters \\

$\boldsymbol{r}$ & received reward\\

$\boldsymbol{s}$ & non-dimensional, normalized observed state\\

$T_x, T_y, T_z$ & thrust in the x, y, and z directions in Newtons\\

$x,y,z$ & coordinates in x, y, and z axis in meters\\

$\dot{x},\dot{y},\dot{z}$ & first derivatives of position in x, y, and z axis in $m/s$ \\

$\ddot{x},\ddot{y},\ddot{z}$ & first derivatives of position in x, y, and z axis in $m/s^2$ \\

$\alpha$ & divergent loss weight\\

$\boldsymbol{\pi}$ & reinforcement learning policy\\

\end{longtable*}}


\section{INTRODUCTION}
\label{sec:intro}

The exponential growth in satellites occupying specific orbital regimes necessitate robust satellite control schemes designed for contested environments. Threats such as space detritus, and non-cooperative spacecraft have long been motivation for satellite path planning algorithms. In previous works, the authors have presented EVADE -- a low size, weight, power and cost (SWaP-C) path planning algorithm for 6 degree-of-freedom (DOF) satellites operating in non-cooperative space \cite{Mehlman2024-az} -- and demonstrated its ability to perform on the edge using an autonomous UAVs \cite{Mehlman2025-yx}. However, like other optimization-based autonomous methods for space, EVADE makes the assumption that at any state there is a safe path that can be computed, and provides no method for incorporating sequential decision making. Thus, in the presence of a strategically proficient adversary where certain actions can lead to `no-win' states, existing path planning algorithms do not provide sufficient behavioral awareness to persistently resist an adversary in pursuit, despite the reality of such a scenario.

Adjacent to spacecraft routing, certain strategic multi-agent scenarios in the space domain such as Pursuit-Evasion have long been of interest to the space community \cite{Weintraub2020-zn}. Traditionally, research in this area focuses on longer range engagement, allowing satellite movement to be described by orbital maneuvers and greatly restricting mobility of agents. However, this approach is not practical for close range engagement which opens the door for a much wider range of behavioral tactics. If we are to produce robust generalizable policies for interacting with non-cooperative agents new approaches to how adversaries are modeled must be developed.

\begin{figure}[thpb]
      \centering
      {\includegraphics[scale=0.5]{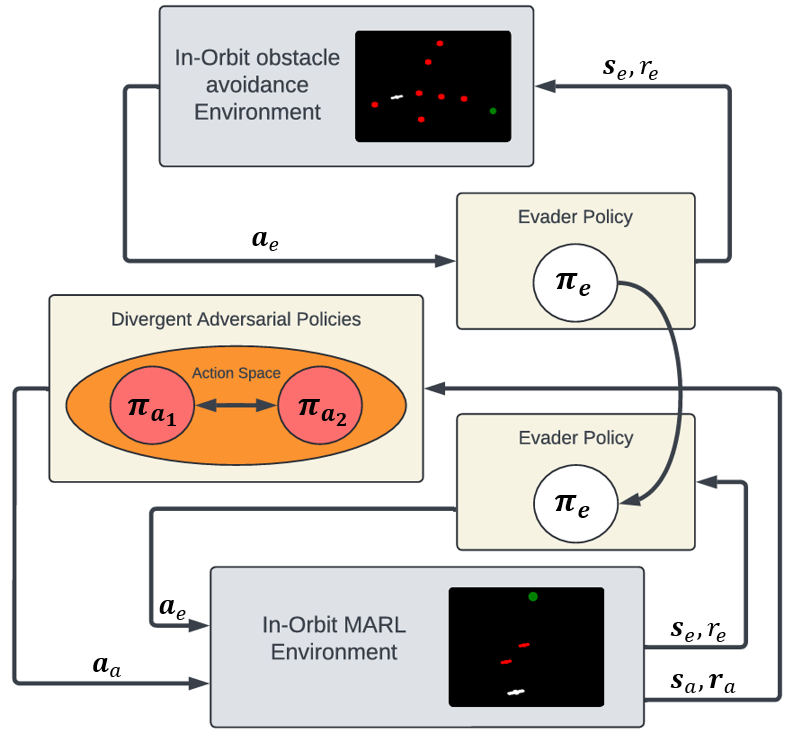}}
      \caption{A depiction of the training scheme we propose. The evader policy $\boldsymbol{\pi}_e$ is initially trained in a static-obstacle avoidance environment. The resulting policy is then retrained in a MARL environment with multiple adversarial policies that are encouraged to produce dissimilar behaviors through a divergent loss term.}
      \vspace{-3mm}
      \label{fig:prompt}
   \end{figure}

The problem statement we explore is an in-orbit cat-and-mouse dynamic where an asset pursues another, which we represent as a rule-based game of capture the flag. An evader spacecraft is tasked with the goal of visiting a randomly chosen area in space, and then returning to its starting point. During this, two adversarial spacecraft attempt to `chase' the evader with an aim to stop it from reaching either goal by blocking or colliding with the spacecraft. We codify the cat-and-mouse dynamic in the capture the flag game because it is a highly complex and strategic task with a wide range of potentially successful adversarial behaviors. Thus, a crucial step to discovering a robust evader `mouse' policy is sufficiently exploring potential adversarial `cat' behaviors.

We propose Divergent Adversarial Reinforcement Learning (DARL): a Multi-Agent Reinforcement Learning (MARL) framework for learning a generalizable evader policy capable of competing with adversarial agents not seen during training. Our multi-staged method leverages several divergent adversarial policies in order to ensure evader exploration of the state space during training for more robust behavior. The evader policy first learns to guide a satellite around stationary obstacles, and is then used to train multiple adversaries in a capture the flag environment before further training on the learned adversaries. An  illustration of our proposed method can be seen in Figure \ref{fig:prompt}. The generalizability of our model is validated by testing in simulation with expert adversarial policies not seen during training. Our key contributions are

\begin{itemize}
    \item We present a novel MARL framework for developing evasive satellite policies for close-proximity multi-agent scenarios.
    \item We present a novel method for learning divergent adversarial policies to be leveraged later in training for improved exploration of an evader policy during training.
    \item We demonstrate the ability of the resulting evader model to outperform several benchmarks, including a satellite path planning algorithm developed in previous works demonstrated on a hardware-in-the-loop testbed environment.
\end{itemize}


\section{RELATED WORKS}
Although efforts in path planning fall short of solving multi-agent games in space, a range of solutions have been proposed in past works mainly focused on solving zero-sum long range Pursuit-Evasion (PE) games. Blasch et. al. proposes a two step orbital maneuver pursuer strategy coupled with a trust-based sensor management policy in order to frame a satellite PE game \cite{Blasch2012-ir}, while Zhang et. al. proposes a dynamic programming approach for computing a Nash Equilibrium \cite{Zhang2024-pm}. Others have employed RL in order to develop optimal policies \cite{Jiang2023-wm,Kartal2021-vr}. However, the PE games proposed often provide data-rich and fully observable environments where agents' actions directly impact the performance of other agents; a considerably less complex task than two-on-one capture the flag. Furthermore, it is common that the general policy for either one or both agents is assumed in order to restrict the optimization problem \cite{Vlahov2018-bn,Zhang2022-nu}, which greatly reduces the ability of solutions to perform in environments where agents act outside of their expected behavior. In our approach we explore methods for multi-agent and partially observable engagement where the strategy of the pursuer may not be known. This requires a more generalizable policy that is not optimized for specific scenarios due to the unpredictable nature of the adversary. 

Developing policies that demonstrate robust behavior for multi-agent scenarios is a complex task, and proven to be a major hurdle in the RL community due to disparities in training and test data \cite{Malik2021-az}. However, recent advances in MARL have shown substantial promise towards developing more generalizable behavior. Qui et. al. propose a method for generalizable MARL by storing a caches of behavioral policies in order to ensure diversity of multi-agent interactions \cite{Qiu2022-ur}. Gupta et. al. provides a different approach by demonstrating the ability to generalize coordination between cooperative MARL agents by improving joint exploration during training \cite{Gupta2021-wc}. While these works have shown that multiple interacting policies can lead to improved exploration and more robust performance, they have not been applied to adversarial scenarios as our work attempts to do. Others have explored adversarial scenarios for learning optimal communications policies, but these problem statements often greatly restrict the potential behavior of an adversary \cite{Sun2023-bd,Rahman2022-yf}. Numerous other MARL methods for zero-sum games such as exploiting known posteriors \cite{Xiong2022-be}, identifying Nash equilibriums \cite{Sayin2021-sc}, and curriculum learning \cite{Chen2023-it} have been proposed in the past, but do not focus on leveraging adversarial interaction for learning generalizable policies. 

Outside of the RL community, Adversarial Training and Generative Adversarial Networks (GANs) are popular methods that leverage adversarial models in order to improve model robustness \cite{Bai2021-by,Creswell2018-te}. While these technologies show significant progress in improving robustness of image classifiers neither have been applied to control of dynamical systems, and both have shown to fall victim to over-fitting \cite{Yazici2020-zh,Yu2022-cy}. However, these methods have proven that adversarial networks possess the ability to produce training data that can be used to improve robustness and general performance of models. In our work, we leverage this assumption to design a training scheme that provides the evader with a more diverse set of experiences during training, thus resulting in more robust final performance.


\section{METHOD}
\label{sec:method}
We demonstrate DARL: a method for leveraging divergent adversarial policies to improve exploration and train robust policies capable of performing in non-cooperative, partially observable, multi-agent environments. We exhibit this by training an evader policy to compete in a zero-sum cat-and-mouse game, similar to capture the flag. In this scenario, an evading `mouse' spacecraft is given a goal point $\sim40$m away, and must visit the goal and return to its initial starting point within the maximum episode length. Competing with the evader spacecraft are two adversarial or `cat' spacecraft, tasked with the goal of stopping the evader from reaching either goal point by colliding with or blocking the evader from reaching either goal. 

The proposed game provides a complex 3 DOF environment with a continuous state space and partial observations for the evader, which cannot be successfully navigated by standard path planning algorithms developed in previous work due to its partial observability and highly dynamic nature. Furthermore, because we do not know a nash-optimal policy for either evader or adversary our MARL framework is entirely exploration based. DARL works in two tiers: first, a base evader policy is trained to reach a desired goal point while avoiding stationary obstacles, second, multiple divergent adversarial policies are learned in a multi-agent cat-and-mouse environment with the base evader policy, which are later leveraged to further train and refine the evader policies. 

\subsection{Evader Problem Definition}

As previously stated, the evader policy must reach a desired goal point and return to its initial position while evading multiple adversaries. Our reward policy is a combination of a continuous reward scheme which is the negative normalized distance to the current goal, and a sparse reward scheme which penalizes the evader in event of collision or episode truncation and rewards the evader if it reaches a goal. The continuous reward allows for the evader to quickly learn to reach its desired goal while the sparse reward allows us to incorporate additional rules from the proposed 'capture the flag' game such as collisions. Initial tuning of the size of the punishments and rewards for the sparse reward scheme was necessary in order to ensure that the finally policy equally prioritized reaching the goal and evading the adversary.

\begin{figure}[thpb]
      \centering
      {\includegraphics[scale=0.8]{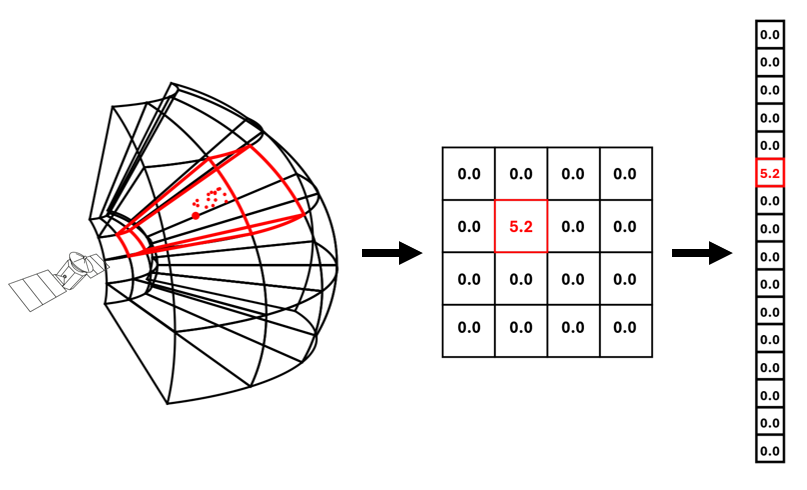}}
      \caption{A description of how the voxelized state space (left) is represented as a flattened matrix $\boldsymbol{H}_f$ (right).}
\vspace{-3mm}
\label{fig:observation}
\end{figure}

In order to emulate the nature of a spacecraft operating in an non-cooperative environment with limited sensing capabilities the evader is assumed to have omnidirectional sensing capabilities within a limited range (10m). Sensor data within this range is represented in a polar histogram which voxelizes the space within a fixed radius from the evader into a series of bins where each bin records the closest distance to any obstacle data point within said bin. The histogram is represented as a flattened $m\times m$ matrix $\boldsymbol{H}_f$ as shown in Figure \ref{fig:observation}. The state of the evader is then described by $\boldsymbol{H_f}$ and the current goal point relative to the evader spacecraft's current position $\boldsymbol{p}_e$ which are concatenated in evaders observation $\boldsymbol{s}_e$:

\begin{align}
\boldsymbol{s} = [\boldsymbol{H}_f,\boldsymbol{g}_e^{E}]
\label{eq:evader_obs}
\end{align}

Additionally, the action space for the evader is comprised of thrusts in the $x, y, z$:

\begin{align}
\boldsymbol{a} = [T_x,T_y,T_z]
\label{eq:action}
\end{align}

The thrusts in any direction is limited to 0.15N. Due to the fact that the evader's observation $\boldsymbol{s}_e$ only provides the state of the environment within a short distance of its location, the environment is partially-observable, and thus the evader policy must learn a Partially Observable Markov Decision Process (POMDP) during training.

\subsection{Adversary Problem Definition}

The adversary is tasked with preventing the evader from reaching either goal point. Thus, its reward policy is a combination of a continuous reward equal to the normalized distance between the evader and its goal, and a sparse reward which includes a punishment if the evader reaches its goal, and a reward if the episode ends early. Each adversary state includes the position of the evader relative to the adversary's position $\boldsymbol{p}_{a_i}$, and the evader's goal position also relative to $\boldsymbol{p}_{a_i}$ which is provided in its observation $\boldsymbol{s}_{ai}$:

\begin{align}
\boldsymbol{s}_{ai} = [\boldsymbol{p}_e^{A_i},\boldsymbol{g}_e^{A_i},\boldsymbol{p}_{a_j}^{A_i}]
\label{eq:adversary_obs}
\end{align}

It should be noted that each adversary is provided a full state of the environment in order to improve adversarial learning and their resulting policies. 

\subsection{Training Scheme}

As mentioned earlier in this section, training occurs in two stages: learning a base evader policy, and learning multiple divergent adversarial policies and a refined evader policy in a multi-agent environment. All policies are trained using Soft Actor-Critic (SAC) with low target entropy \cite{haarnoja2018softactorcriticoffpolicymaximum}. An off-policy RL algorithm was chosen because off-policy algorithms possess separate actor networks. This allows us to influence the state-action pairs sampled during training without directly manipulating the policy learning algorithm when encouraging divergent adversarial behavior.

\begin{figure}[thpb]
      \centering
      {\includegraphics[scale=1.2]{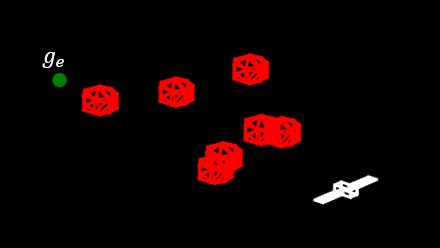}}
      \caption{An image of the obstacle evasion environment used to train the base evader policy, where the evader (white), must reach $g_e$, without colliding with any of the obstacles (red).}
\vspace{-3mm}
\label{fig:evadeobs}
\end{figure}

\subsubsection{Stage I: Learning a Base Evader Policy}

The base evader is trained on the same single-agent obstacle avoidance environment EVADE was developed on. The evader is tasked with reaching a goal point approximately $40m$ in the distance with several randomly initialized obstacles in its path. The evader receives the same observation and follows the same reward scheme as previously defined. However, the base evader policy is only tasked with reaching a single goal point. In order to train a successful policy, a curriculum learning strategy is employed by increasing the number of obstacles initialized at the beginning of each episode as training progresses. The number of obstacles initialized begins with 1, and increases linearly to a predetermined number before training ends. An image of an episode starting with 7 obstacles initialized can be seen in Figure \ref{fig:evadeobs}.

\subsubsection{Stage II: Learning from Divergent Adversaries}

During Stage II we propose a multi-stage training regiment in which we train both multiple divergent adversarial policies and a refined evader policy in a cat and mouse environment. First, multiple adversarial policies learn to prevent the base evader policy from reaching either goal. Then, after a predetermined number of training steps the base evader policy begins retraining in the same environment in order to learn to evade the adversarial policies. Once the evader begins retraining, the adversaries stop network updates to prevent non-stationarity of the environment. The final evader policy demonstrates improved generalizable behavior by performing in an environment with expert adversarial policies not seen during training.

In order to ensure that the evader learns a robust and generalizable policy, multiple adversarial policies are learned. However, this does not mitigate how the adversarial policies are prone to converging upon sub-optimal policies which do not challenge the evader, leading to over-fitting and a lack of robustness. In order to combat this, we implement a divergent loss term which encourages each actor to explore different subspaces of the action space, and prevents adversaries from converging upon similar sub-optimal policies early-on during training. In turn, this promotes exploration of the evader and leads to more robust behavior. 

The divergent loss term implemented $\mathcal{L}_{KL}$ is the weighted average MSE between the Kullback–Leibler divergence of the actions taken from two networks given the same observations and a predetermined constant $c_{KL}$.

\begin{algorithm}[t]
\caption{Learning Divergent Adversarial Policies}
\label{algo:div}
\begin{algorithmic}[1]
\State Initialize adversarial policies $\boldsymbol{\pi}_{\theta_a}= [\pi_{\theta_{a_0}},...,\pi_{\theta_{a_n}}]$, $\boldsymbol{Q}{\phi_a}=[Q_{\phi_{a_0}},...,Q_{\phi_{a_n}}]$ and off-policy RL algorithm with replay buffers $\boldsymbol{\mathcal{D}}_{a}=[\mathcal{D}_{a_0},...\mathcal{D}_{a_n}]$
\State Initialize evader policy $\pi_{\theta_{e}}$, $Q_{\phi_{e}}$ and off-policy RL algorithm with replay buffer $\mathcal{D}_{e}$
\For{each iteration}
\If{Episode reset}
\State pick random policy from $\boldsymbol{\pi}_{\theta_a}$,$\boldsymbol{Q}_{\phi_a}$
\EndIf
\State execute sampled actions and record observations
\If{update adversary}
\For{policy $\boldsymbol{\pi}_{\theta_{a_i}}$ in $\boldsymbol{\pi}_{\theta_a}$}
\State sample batch $B=\{\boldsymbol{s},\boldsymbol{a},r,\boldsymbol{s}',d\}$ from $\mathcal{D}_{a_i}$
\State compute $\pi_{\theta_{a_i}}(\cdot | \boldsymbol{s})$
\For{policy $\boldsymbol{\pi}_{\theta_{a_j}}$ in $\boldsymbol{\pi}_{\theta_a}$}
\If{$j \neq i$}
\State compute $\pi_{\theta_{a_j}}(\cdot | \boldsymbol{s})$
\EndIf
\EndFor
\State $\theta \leftarrow \theta - \nabla_{\theta} \mathcal{L}_{RL}(\theta)$ 
\State $\theta \leftarrow \theta  -  \nabla_{\theta} \mathcal{L}_{KL_i}(\theta)$ 
\EndFor
\EndIf
\EndFor
\end{algorithmic}
\end{algorithm}

\begin{align}
\mathcal{L}_{KL_i} = 
\frac{\alpha}{n-1}
\sum_{\substack{j=0,\\j\neq i}}^{n-1}
\big(c_{KL} - 
D_{KL}\big(\pi_{\theta_{a_i}}(\cdot | \boldsymbol{s}_{a_i})||\pi_{\theta_{a_j}}(\cdot | \boldsymbol{s}_{a_i})\big)
\big)^2
\label{eq:kl_loss}
\end{align}

where $n$ is the number of divergent adversarial policies, and $\alpha$ is a predetermined weight which follows a linear decay schedule during training:

\begin{align}
\alpha = 
\left\{
\begin{array}{lr}
1 & t < t_{start}\\
 1 - \frac{t-t_{start}}{t_{stop}-t_{start}} & t_{stop} <t < t_{start}\\
0 & t > t_{stop} \\
\end{array}
\right.
\end{align}

where $t$ is the current timestep, and $t_{start},t_{stop}$ are the start and stop cutoffs for implementing divergent loss decay. The parameters of the adversary's actor network are then updated based off of $\mathcal{L}_{KL}$ in addition to their standard loss term $\mathcal{L}_{RL}$. 

Algorithm \ref{algo:div} describes our divergent adversarial learning algorithm where there are two main differences between our method, and a baseline MARL algorithm. The first is lines 4 and 5, where at the beginning of each episode we randomly assign one of the $n$ adversarial models to both of the two adversarial agents in the environment. The second is in lines 8 through 15, where we compute $\mathcal{L}_{KL_i}$ for each adversarial model in order to ensure dissimilar policies. The hyper parameters used during training can be found in the Appendix.

Additionally, encouraging divergence between models can cause exploding gradients early-on in training; hindering the adversary's ability to learn. In order to prevent this, the output layer of the actor network must be bounded by a non-monotonic activation function. Thus, we find a periodic activation function is necessary in order to allow the network to optimize both loss parameters.


\section{EXPERIMENT AND RESULTS}

\subsection{Experimental Setup}

In simulation, the dynamics of the evader and adversary spacecrafts can be described using the Clohessy Wiltshire (CW) Equations, which model the behavior of a spacecraft in relation to a single point moving on a near circular orbit:
\vspace{-2mm}
\begin{align}
\ddot{x} = 3n^2x + 2n\dot{y} + \frac{T_x}{m}
\label{eq:cwx}
\end{align}
\vspace{-4mm}
\begin{align}
\ddot{y} = -2n\dot{x} + \frac{T_y}{m}
\label{eq:cwy}
\end{align}
\vspace{-4mm}
\begin{align}
\ddot{z} = -n^2z + \frac{T_z}{m}
\label{eq:cwz}
\end{align}

Where $x,y,$ and $z$ are the Cartesian coordinates of the spacecraft relative to the single moving point, which is treated as the origin of out localized coordinate frame.

The cat and mouse training prompt is described in previous sections. During testing, both adversary and evader spacecrafts are given the same physical parameters (mass, inertia, etc.). Adversaries are modeled as a ball with a $1m$ radius, and represented as a point cloud. The evader must successfully come within $3m$ of a goal point, and return to its starting position without coming within $1m$ of an adversary for the episode to be considered successful. If the evader is capable of reaching the first goal but not returning to its starting position the episode is considered a partial success. And, if the evader does not reach either goal the episode is considered a failure.

In order to provide an impartial and standardized comparison of model performance, all models are tested in an environment with hand designed heuristic adversarial policies which receive expert knowledge of the nature of the cat and mouse game. Two expert adversarial policies are implemented, one that attempts to collide with the evader, and a second that attempts to block any path from the evader to the goal point. In every test episode, one of each heuristic policy is designated to a single adversary. The heuristic policy implemented is shown in Algorithm \ref{algo:adv}, where $h()$ is a function that determines the point the adversary should travel to, and $c()$ computes the desired action input.  

\begin{algorithm}[t]
\caption{Adversary Heuristic Policy}
\label{algo:adv}
\begin{algorithmic}[1]
\State Initialize adversary at $\boldsymbol{p}_a$
\For{every timestep}
\State collect observation $\boldsymbol{s}_a$
\For{every $N$ steps}
\State $\boldsymbol{g}_a \leftarrow h(\boldsymbol{s}_a)$
\EndFor
\State $\boldsymbol{a}_a \leftarrow c(\boldsymbol{g}_a,\boldsymbol{s}_a)$
\State take action $\boldsymbol{a}_a$
\EndFor
\end{algorithmic}
\end{algorithm}

\begin{figure}[thpb]
      \centering
      {\includegraphics[scale=0.7]{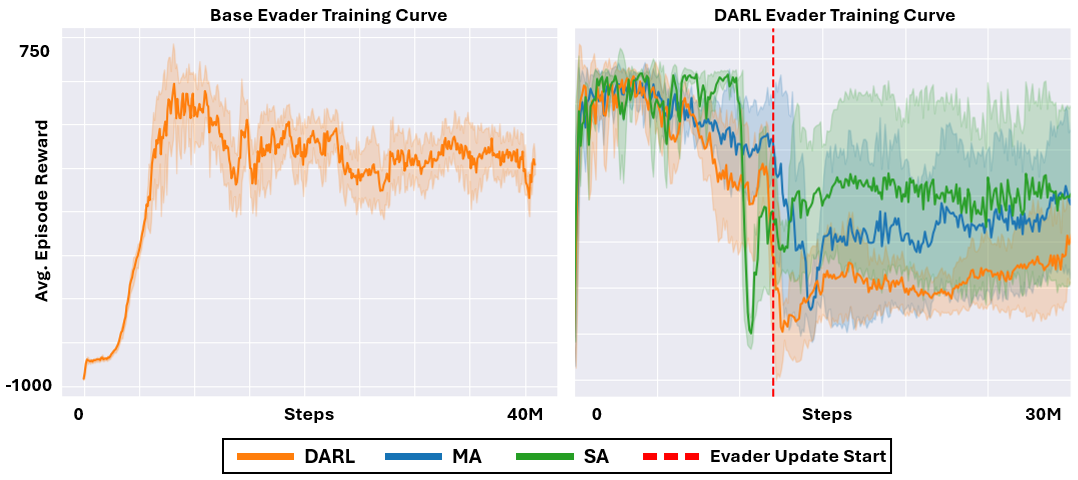}}
      \caption{Base evader training curve (left), and DARL, MA, and BA Evader training curves (right). The dotted red line marks the beginning of the evader policy's network updates.}
\label{fig:be_curve}
\end{figure}

\subsection{Evaluation Benchmarks}
\label{subsec:eval}

Several methods are used as benchmarks in order to demonstrate our methods robustness, and ability to successfully evade intelligent adversaries, the several benchmarks are listed below. 

\begin{enumerate}
    \item \textit{EVADE}: EVADE is a optimization based path planner designed specifically for spacecraft operating in non-cooperative space environments with passive obstruction \cite{Mehlman2024-az}.

    \item \textit{Base Evader Policy (BE)}: This is the resulting policy of Stage I of training before it is retrained in Stage II.

    \item \textit{Single Adversary (SA)}: This method entails a single adversarial policy, thus no $\mathcal{L}_{KL}$ term and a standard linear activation.

    \item \textit{Base Multi-Adversary (MA)}: This method entails multiple adversarial policies, however, uses a standard linear activation and no $\mathcal{L}_{KL}$ term.

    \item \textit{DARL}: This is our full method discussed in Section \ref{sec:method}. The model is trained with 2 divergent adversarial policies.

    \item \textit{No Sine Activation (NSA)}: As a brief ablation demonstration, this implementation of our method uses a standard linear activation layer for the actor network.
\end{enumerate}

Methods 3-6 are all trained using the two stage training scheme discussed in Section \ref{sec:method}. Methods which include multiple adversaries entail training 2 separate policies.

\subsection{Results and Discussion}

\begin{table}
    \centering
    \caption{Average performance of models trained using benchmark methods listed in Section \ref{subsec:eval}.}
    \ra{1.1}
    \begin{tabular}{ccc}
        \midrule
        \midrule
         Method & Failure (\%) & Success (\%)\\
         \midrule
         EVADE & $100\%$ &  $0\%$ \\
         BE &  $58.7\%$ & $7.32\%$ \\
         SA & $40\%$  & $26\%$\\
         MA & $35.5\%$  & $36.5\%$\\
         \textbf{DARL} & $\mathbf{29\%}$  & $\mathbf{48.5}\%$\\
         NSA & $32\%$  & $40\%$\\
         \midrule
         \midrule
    \end{tabular}
    \label{tab:res}
\end{table}

Training curves for the baseline, DARL, MA, and SA evader can be seen in Figure \ref{fig:be_curve}. In the figure, the evader update start line is placed at 12M timesteps, which was chosen in order to provide the adversarial policies with sufficient training to converge on a policy. Table \ref{tab:res} shows the test performance of each method where a success is defined by reaching the goal and returning to its initial position and a failure is defined by the spacecraft not being able to reach the initial goal. In Table \ref{tab:res} it can be clearly seen that DARL outperforms all benchmarks both in maximizing successful episodes, and minimizing failures. It should also be noted that our two staged method clearly demonstrates the ability to produce policies for satellite autonomy that can perform in highly contested environments that standard satellite path-planning algorithms (EVADE) fail in. Although EVADE and the base evader are intended for partially-observable non-cooperative environments, neither were designed to protect a spacecraft in a strategic multi-agent game. Finally, in Table \ref{tab:res} it can be seen that the top 3 performing models all involve training on multiple adversarial models, emphasizing the value gained from training in environments that force the learning policy to widen their exploration of the state space. From these results, it is a reasonable assumption that in environments with larger state spaces, or more potential adversarial behaviors a greater number of divergent policies would only improve exploration during training and result in more robust performance of the final model. However, as the number of divergent policies increase there should be diminishing returns once a number of policies that can sufficiently explore all potential adversarial behavior has been reached, which should be explored in future work. It is also important to note that with each change in the number of divergent adversarial policies the divergent loss constant $c_{KL}$ and train steps for the adversary must be re-tuned.

\begin{figure}[thpb]
      \centering
      {\includegraphics[scale=0.6]{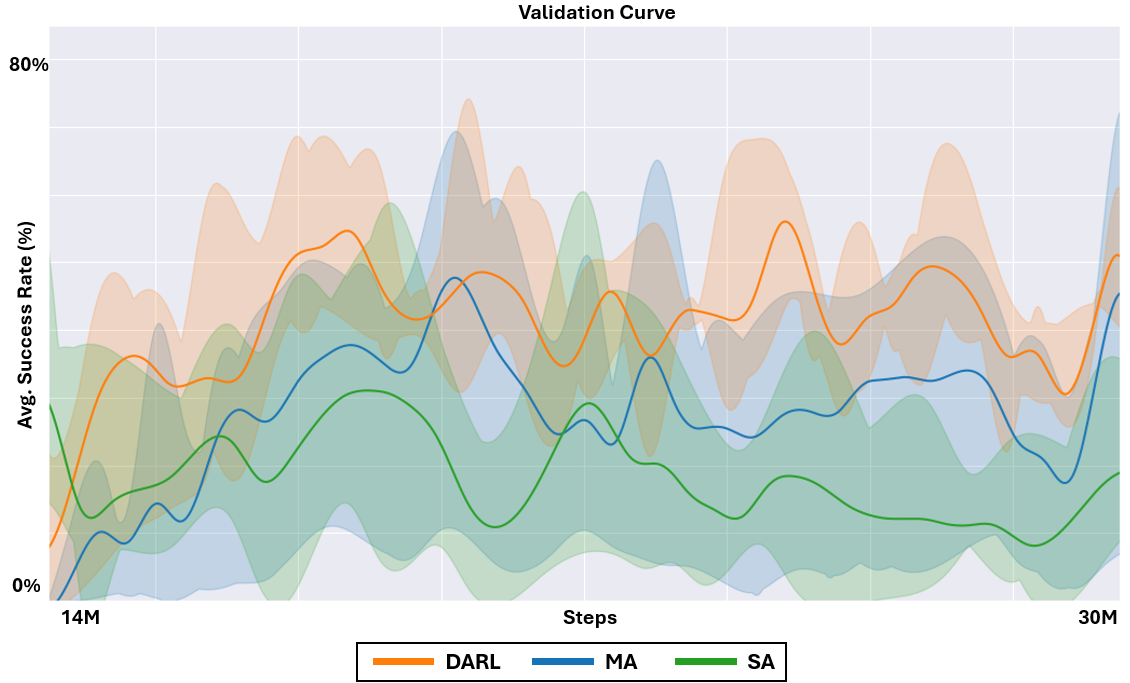}}
      \caption{Validation curve comparing the average test performance of the DARL, MA, and SA models trained at different timesteps.}
\label{fig:val_curve}
\end{figure}

In addition to comparing the testing performance of the models trained, Figure \ref{fig:val_curve} shows the average test performance of several methods throughout training. The Figure compares the average test performance of the policies trained through the Single-Adversary (SA), Multi-Adversary (MA), and DARL methods starting once the evader model begins updates. It should be noted that a base evader trained on multiple divergent adversaries both learns faster, and consistently learns more robust behavior for cat and mouse games. 

Aside from general performance of our method in comparison to baselines, it is worth noting that the adversaries during training and testing receive full state information of the environment which may not be realistic of a real-world scenario. Due to the generalizable properties our trained policy has shown, as well as the partial-observability of the evader which makes it difficult to interpret adversarial behavior outside of a short range it is reasonable to assume that a degradation in adversary performance would not have an negative impact on evader performance. However, it is difficult to predict model performance when operating in environment states not seen during training, and thus this should be explored in future work.

\subsection{Sim-to-real Feasibility and Applications}

Although we do not attempt sim-to-real transfer, we believe with a series of adjustments to our simulation environment the resulting DARL models can be integrated into our UAV testbed via the MAGPIE System which is designed to emulate spacecraft architecture and behavior \cite{Mehlman2025-yx}. Primarily, this would entail modifying the environment to emulate UAV that take positional commands rather than spacecraft. Although training a model to provide positional commands rather than force control would simplify the problem at hand, it would allow for integration of additional safety features often employed in UAV sim-to-real transfer \cite{Ugurlu2022-jm}. Additionally, as shown by previous works with sufficient domain randomization successful sim-to-real on UAVs is feasible \cite{Loquercio2020-dz}.

Outside of hardware test-beds, RL policies have yet to be implemented on spacecraft in operational environments. However, space provides an appealing setting for autonomous applications due to communications-denied environments, and risks inherent to human-in-the-loop spaceflight operations \cite{Tipaldi2022-ey}. Furthermore, modern satellites posses computing power capable of running standard RL models, which are generally smaller than models trained using other ML methods. Reduced launch cost and the associated increase in launch frequency of small satellites provides potential for running DARL on a satellite in coming year, an effort our team is actively seeking with potential sponsors. Future work concerning sim-to-real should also include developing safety measures and assurances around the cat-and-mouse interaction.


\section{CONCLUSION}
Our work proposes a new method for leveraging multiple divergent adversarial policies in order to train a robust and generalizable RL policy for satellite control in non-cooperative multi-agent environments known as DARL. We demonstrate this method by training an RL model to compete in a partially-observable game of capture the flag. The game is meant to emulate a multi agent cat-and-mouse satellite dynamic with 2 intelligent adversarial `cat' agents pursuing an evading `mouse' agent. During testing, we demonstrate not only that training with diverse adversarial policies consistently leads to more robust behavior, but also highlight the critical need for POMDP-based policies by showing the failure of standard satellite routing algorithms to safeguard spacecraft in complex, strategic multi-agent environments. DARL, deployed on a spacecraft, will enable the successful evasion of adversarial space vehicles so that the mouse can complete its mission.

\section{APPENDIX}
\label{sec:appendix}
In this section we provide the hyperparameters used to train both the base evader policy in Stage I, the final evader in stage II, and divergent adversarial policies. All training was done using RLlib which is an open-sourced RL library developed by Ray. The SAC hyperparameters used during both Stage I and II were the same, and thus provided in Table \ref{tab:sac} separate from Stage I and II hyperparameters (Table \ref{tab:paramsI} and Table \ref{tab:paramsII} respectively). Any hyperparameter not described in Table \ref{tab:sac} was set to RLlib's default.

\begin{table}[h!]
\centering
\caption{SAC hyperparameters}
\begin{tabular}{lcp{8cm}}
\hline
\hline
\textbf{Hyperparameter} & \textbf{Value} & \textbf{Description} \\
\hline
$\gamma$       & 0.99       & Decay factor \\
Batch Size     & 256        & Train batch size \\
Replay Buffer  & 1e7        & Train replay buffer size \\
Target Entropy & 0.3        & SAC target entropy \\
Train Intensity& 3          & SAC train intensity \\
Gradient Clip  & 10         & Gradient clipping maximum value \\
Actor Architecture & 512x512& Actor network hidden layers architecture \\
Critic Architecture& 512x512& Critic network hidden layers architecture \\
\hline
\hline
\end{tabular}
\label{tab:sac}
\end{table}

\begin{table}[h!]
\centering
\caption{Stage I unique hyperparameters for training evader base policy}
\begin{tabular}{lcp{8cm}}
\hline
\hline
\textbf{Hyperparameter} & \textbf{Value} & \textbf{Description} \\
\hline
$t_{train}$    & 30e7       & Total train timesteps  \\
$o_{max}$      & 5        & Maximum number of obstacles at end of training  \\
$T_{max}$      & 0.15       & Maximum thrust in x,y, or z (Newtons) \\
\hline
\hline
\end{tabular}

\label{tab:paramsI}
\end{table}

\begin{table}[h!]
\centering
\caption{Stage II hyperparameters for training evader and adversary policies}
\begin{tabular}{lcp{8cm}}
\hline
\hline
\textbf{Hyperparameter} & \textbf{Value} & \textbf{Description} \\
\hline
$c_{KL}$       & 1.0        & KL divergence loss constant  \\
$\alpha$       & 100        & KL divergence loss scale term  \\
$t_{train}$    & 30e7       & Total train timesteps  \\
$t_{start}$    & 0          & KL divergence decay start  \\
$t_{stop}$     & 10e7       & KL divergence implementation end  \\
$t_{update}$   & 12e7       & Evader update start timestep  \\
$n$            & 2          & Number of divergent adversarial policies \\
$T_{max}$      & 0.15       & Maximum thrust in x,y, or z (Newtons) \\
\hline
\hline
\end{tabular}
\label{tab:paramsII}
\end{table}

\pagebreak


\section*{Acknowledgments}
The authors thank Joseph Abramov and Murphy Klein for their contributions to this work.

\bibliography{references}

@ARTICLE{Tipaldi2022-ey,
  title        = {Reinforcement learning in spacecraft control applications:
                  Advances, prospects, and challenges},
  author       = {Tipaldi, Massimo and Iervolino, Raffaele and Massenio, Paolo
                  Roberto},
  journal = {Annual Reviews in Control},
  volume       = {54},
  pages        = {1--23},
  year         = {2022},
  doi          = {10.1016/j.arcontrol.2022.07.004}
}

@ARTICLE{Loquercio2020-dz,
  title        = {Deep Drone Racing: From Simulation to Reality With Domain
                  Randomization},
  author       = {Loquercio, Antonio and Kaufmann, Elia and Ranftl, René and
                  Dosovitskiy, Alexey and Koltun, Vladlen and Scaramuzza, Davide},
  journal = {IEEE Transactions on Robotics},
  volume       = {36},
  issue        = {1},
  pages        = {1--14},
  year         = {2020},
  doi          = {10.1109/TRO.2019.2942989}
}

@ARTICLE{Ugurlu2022-jm,
  title        = {Sim-to-Real Deep Reinforcement Learning for Safe End-to-End
                  Planning of Aerial Robots},
  author       = {Ugurlu, Halil Ibrahim and Pham, Xuan Huy and Kayacan, Erdal},
  journal = {Robotics},
  volume       = {11},
  issue        = {5},
  year         = {2022},
  doi          = {10.3390/robotics11050109}
}

@MISC{Yu2022-cy,
  title        = {Understanding Robust Overfitting of Adversarial Training and
                  Beyond},
  author       = {Yu, Chaojian and Han, Bo and Shen, Li and Yu, Jun and Gong,
                  Chen and Gong, Mingming and Liu, Tongliang},
  journaltitle = {arXiv [cs.LG]},
  year         = {2022},
  eprintclass  = {cs.LG},
  doi          = {10.48550/arXiv.2206.08675}
}

@MISC{Yazici2020-zh,
  title        = {Empirical Analysis of Overfitting and Mode Drop in {GAN}
                  Training},
  author       = {Yazici, Yasin and Foo, Chuan-Sheng and Winkler, Stefan and
                  Yap, Kim-Hui and Chandrasekhar, Vijay},
  journaltitle = {arXiv [cs.LG]},
  year         = {2020},
  eprintclass  = {cs.LG},
  doi          = {10.1109/ICIP40778.2020.9191083}
}

@ARTICLE{Creswell2018-te,
  title        = {Generative Adversarial Networks: An Overview},
  author       = {Creswell, Antonia and White, Tom and Dumoulin, Vincent and
                  Arulkumaran, Kai and Sengupta, Biswa and Bharath, Anil A},
  journal = {IEEE Signal Processing Magazine},
  volume       = {35},
  issue        = {1},
  pages        = {53--65},
  year         = {2018},
  doi          = {10.1109/MSP.2017.2765202}
}

@MISC{Bai2021-by,
  title        = {Recent Advances in Adversarial Training for Adversarial
                  Robustness},
  author       = {Bai, Tao and Luo, Jinqi and Zhao, Jun and Wen, Bihan and Wang,
                  Qian},
  journaltitle = {arXiv [cs.LG]},
  year         = {2021},
  eprintclass  = {cs.LG},
  doi          = {10.24963/ijcai.2021/591}
}

@MISC{Chen2023-it,
  title        = {Accelerate Multi-Agent Reinforcement Learning in Zero-Sum
                  Games with Subgame Curriculum Learning},
  author       = {Chen, Jiayu and Xu, Zelai and Li, Yunfei and Yu, Chao and
                  Song, Jiaming and Yang, Huazhong and Fang, Fei and Wang, Yu
                  and Wu, Yi},
  journaltitle = {arXiv [cs.LG]},
  year         = {2023},
  eprintclass  = {cs.LG},
  doi          = {10.1609/aaai.v38i10.29011}
}

@MISC{Sayin2021-sc,
  title        = {Decentralized {Q}-Learning in Zero-sum Markov Games},
  author       = {Sayin, Muhammed O and Zhang, Kaiqing and Leslie, David S and
                  Basar, Tamer and Ozdaglar, Asuman},
  journaltitle = {arXiv [cs.GT]},
  year         = {2021},
  eprintclass  = {cs.GT},
  doi          = {10.48550/arXiv.2106.02748}
}

@MISC{Xiong2022-be,
  title        = {A Self-Play Posterior Sampling Algorithm for Zero-Sum Markov
                  Games},
  author       = {Xiong, Wei and Zhong, Han and Shi, Chengshuai and Shen, Cong
                  and Zhang, Tong},
  journaltitle = {arXiv [cs.LG]},
  year         = {2022},
  eprintclass  = {cs.LG},
  doi          = {10.48550/arXiv.2210.01907}
}

@MISC{Rahman2022-yf,
  title        = {{AdverSAR}: Adversarial Search and Rescue via Multi-Agent
                  Reinforcement Learning},
  author       = {Rahman, Aowabin and Bhattacharya, Arnab and Ramachandran,
                  Thiagarajan and Mukherjee, Sayak and Sharma, Himanshu and
                  Fujimoto, Ted and Chatterjee, Samrat},
  journaltitle = {arXiv [cs.RO]},
  year         = {2022},
  eprintclass  = {cs.RO},
  doi          = {10.48550/arXiv.2212.10064}
}

@INPROCEEDINGS{Sun2023-bd,
  title     = {Certifiably Robust Policy Learning against Adversarial
               Multi-Agent Communication},
  author    = {Sun, Yanchao and Zheng, Ruijie and Hassanzadeh, Parisa and Liang,
               Yongyuan and Feizi, Soheil and Ganesh, Sumitra and Huang, Furong},
  booktitle = {The Eleventh International Conference on Learning Representations},
  year      = {2023},
  pages     = {0-10},
  url       = {https://openreview.net/forum?id=dCOL0inGl3e}
}

@MISC{Gupta2021-wc,
  title        = {{UneVEn}: Universal Value Exploration for Multi-Agent
                  Reinforcement Learning},
  author       = {Gupta, Tarun and Mahajan, Anuj and Peng, Bei and Böhmer,
                  Wendelin and Whiteson, Shimon},
  journaltitle = {arXiv [cs.LG]},
  year         = {2021},
  eprintclass  = {cs.LG},
  doi          = {10.48550/arXiv.2010.02974}
}

@MISC{Qiu2022-ur,
  title        = {{RPM}: Generalizable Behaviors for Multi-Agent Reinforcement
                  Learning},
  author       = {Qiu, Wei and Ma, Xiao and An, Bo and Obraztsova, Svetlana and
                  Yan, Shuicheng and Xu, Zhongwen},
  journaltitle = {arXiv [cs.MA]},
  year         = {2022},
  eprintclass  = {cs.MA},
  doi          = {10.48550/arXiv.2210.09646}
}

@MISC{Malik2021-az,
  title        = {When Is Generalizable Reinforcement Learning Tractable?},
  author       = {Malik, Dhruv and Li, Yuanzhi and Ravikumar, Pradeep},
  journaltitle = {arXiv [cs.LG]},
  year         = {2021},
  eprintclass  = {cs.LG},
  doi          = {10.48550/arXiv.2101.00300}
}

@ARTICLE{Kartal2021-vr,
  title        = {Optimal game theoretic solution of the pursuit-evasion intercept problem using on-policy reinforcement learning},
  author       = {Kartal, Yusuf and Subbarao, Kamesh and Dogan, Atilla and Lewis, Frank},
  journal = {International Journal of Robust and Nonlinear Control},
  volume       = {31},
  issue        = {16},
  pages        = {7886--7903},
  year         = {2021},
  doi          = {10.1002/rnc.5719}
}

@INPROCEEDINGS{Vlahov2018-bn,
  title     = {On Developing a {UAV} Pursuit-Evasion Policy Using Reinforcement Learning},
  author    = {Vlahov, Bogdan and Squires, Eric and Strickland, Laura and Pippin, Charles},
  booktitle = {2018 17th IEEE International Conference on Machine Learning and Applications (ICMLA)},
  pages     = {859--864},
  year      = {2018},
  doi       = {10.1109/ICMLA.2018.00138}
}

@ARTICLE{Zhang2022-nu,
  title        = {Near-optimal interception strategy for orbital pursuit-evasion using deep reinforcement learning},
  author       = {Zhang, Jingrui and Zhang, Kunpeng and Zhang, Yao and Shi, Heng and Tang, Liang and Li, Mou},
  journal = {Acta Astronautica},
  volume       = {198},
  pages        = {9--25},
  year         = {2022},
  doi          = {10.1016/j.actaastro.2022.05.057}
}

@ARTICLE{Jiang2023-wm,
  title        = {Orbital Interception Pursuit Strategy for Random Evasion Using Deep Reinforcement Learning},
  author       = {Jiang, Rui and Ye, Dong and Xiao, Yan and Sun, Zhaowei and Zhang, Zeming},
  journal = {Space: Science \& Technology},
  volume       = {3},
  pages        = {0086},
  year         = {2023},
  doi          = {10.34133/space.0086}
}

@ARTICLE{Zhang2024-pm,
  title        = {Fixed-Time Zero-Sum {Pursuit–Evasion} Game Control of Multisatellite via Adaptive Dynamic Programming},
  author       = {Zhang, Zhixuan and Zhang, Kun and Xie, Xiangpeng and Sun, Jiayue},
  journal = {IEEE Transactions on Aerospace and Electronic Systems},
  volume       = {60},
  issue        = {2},
  year         = {2024},
  pages        = {2224-2235},
  doi          = {10.1109/TAES.2024.3351810}
}

@INPROCEEDINGS{Blasch2012-ir,
  title     = {Orbital satellite pursuit-evasion game-theoretical control},
  author    = {Blasch, Erik P and Pham, Khanh and Shen, Dan},
  booktitle = {2012 11th International Conference on Information Science, Signal
               Processing and their Applications (ISSPA)},
  year      = {2012},
  pages     = {1007-1012},
  doi       = {10.1109/ISSPA.2012.6310436}
}

@INPROCEEDINGS{Weintraub2020-zn,
  title     = {An Introduction to Pursuit-evasion Differential Games},
  author    = {Weintraub, Isaac E and Pachter, Meir and Garcia, Eloy},
  booktitle = {2020 American Control Conference (ACC)},
  pages     = {1049--1066},
  year      = {2020},
  doi       = {10.48550/arXiv.2003.05013},
}

@INPROCEEDINGS{Mehlman2025-yx,
  title     = {The {MAGPIE}: Satellite Autonomy for Uncooperative Environments},
  author    = {Mehlman, Cameron and {Kounios} and {Lai} and {Prasad} and {Brown}
               and {Hughes} and {Chalamalasetty} and {Distler} and {Dilone} and
               {Palomino} and {Goel} and Falco, Gregory},
  booktitle = {Hawaii International Conference on System Sciences 2025
               (HICSS-57)},
  pages     = {7275-7285},
  year      = {2025},
  url       = {https://hdl.handle.net/10125/109722}
}

@INPROCEEDINGS{Mehlman2024-az,
  title     = {An Autonomous Satellite Collision Avoidance and Adversary Evasion
               Path Planning Algorithm for the Space Environment},
  author    = {Mehlman, Cameron and Falco, Gregory},
  booktitle = {2024 American Control Conference (ACC)},
  pages     = {3055--3061},
  year      = {2024},
  doi       = {10.23919/ACC60939.2024.10644335}
}

@misc{haarnoja2018softactorcriticoffpolicymaximum,
      title={Soft Actor-Critic: Off-Policy Maximum Entropy Deep Reinforcement Learning with a Stochastic Actor}, 
      author={Tuomas Haarnoja and Aurick Zhou and Pieter Abbeel and Sergey Levine},
      year={2018},
      eprint={1801.01290},
      archivePrefix={arXiv},
      primaryClass={cs.LG},
      url={https://arxiv.org/abs/1801.01290}, 
}

\end{document}